\documentclass[a4paper]{article}

\usepackage{INTERSPEECH2018}
\usepackage{adjustbox}
\usepackage{amsmath,amsfonts,graphicx,hyperref,xcolor}
\usepackage{mathtools,upquote}
\usepackage{bbm}
\usepackage{amsthm}
\usepackage[caption = false]{subfig}
\usepackage{multirow}
\usepackage{array}
\usepackage{booktabs}
\usepackage{adjustbox}
\usepackage{float}
\usepackage{enumitem}
\usepackage{verbatim}
\usepackage{pbox}
\usepackage[symbol]{footmisc}

\usepackage{algorithm}
\usepackage{algpseudocode}

\usepackage[usestackEOL]{stackengine}
\setstackgap{S}{2pt}

\algnewcommand{\IIf}[1]{\State\algorithmicif\ #1\ \algorithmicthen}
\algnewcommand{\EndIIf}{\unskip\ \algorithmicend\ \algorithmicif}

\DeclareMathOperator{\modm}{\text{ mod }}
\newcommand{\prob}[2]{\underset{#1} {\bf{Pr}}\left[ #2 \right]}

\newcommand{\numentries}{n}
\newcommand{\hashbuckets}{m}
\newcommand{\stringlength}{s}

\newcommand{\weightsize}{w}
\newcommand{\numbins}{k}
\newcommand{\featureset}{S}
\newcommand{\featureuniverse}{U}
\newcommand{\featuresubset}{S'}

\title{Statistical Model Compression for Small-Footprint Natural Language Understanding}
\name{Grant P. Strimel\textsuperscript{*}\thanks{\textsuperscript{*}The authors have equal contribution to this work. The names are in alphabetical order.}, Kanthashree Mysore Sathyendra\textsuperscript{*}\footnotemark[1], Stanislav Peshterliev\textsuperscript{*}\footnotemark[1]}
\address{
  Alexa Machine Learning, Amazon.com}
\email{\{gsstrime,ksathyen,stanislp\}@amazon.com}

\begin{document}

\maketitle
\begin{abstract}
In this paper we investigate statistical model compression applied to natural language understanding (NLU) models. Small-footprint NLU models are important for enabling offline systems on hardware restricted devices, and for decreasing on-demand model loading latency in cloud-based systems. To compress NLU models, we present two main techniques, parameter quantization and perfect feature hashing. These techniques are complementary to existing model pruning strategies such as L1 regularization. We performed experiments on a large scale NLU system. The results show that our approach achieves 14-fold reduction in memory usage compared to the original models with minimal predictive performance impact.






\end{abstract}

\section{Introduction}

Voice-assistants with natural language understanding (NLU) \cite{sarikaya2017technology}, such as Amazon Alexa, Apple Siri, Google Assistant, and Microsoft Cortana, are increasing in popularity. However with their popularity, there is a growing demand to support availability in many contexts and wide range of functionality.

To support Alexa in contexts with no internet connection, Panasonic and Amazon announced a partnership to bring offline voice control services to car navigation commands, temperature control, and music playback \cite{Locklear2018}. These services are ``offline'' because the local system running the voice-assistant may not have an internet access. Thus, instead of sending the user's request for cloud-based processing, everything including NLU has to be performed locally on a hardware restricted device. However, cloud-based NLU models have large memory footprints which make them unsuitable for local system deployment without appropriate compression.

Furthermore, to support wide range of functionality, Amazon Alexa and Google Assistant support skills built by external developers. Each skill has NLU models that extend the functionality of the main NLU models. Since there are many skills, their NLU models are loaded on demand only when needed to process user request \cite{kumar2017just}. If the skill NLU model sizes are large, loading them into memory adds significant latency to utterance recognition. Thus, small-footprint NLU models are important for providing quick NLU response and good customer experience. 

Typically NLU models consist of domain classification (DC), intent classification (IC) and named-entity recognition (NER) models. DC predicts the general domain class of a user utterance such Music, Shopping, and Cinema. IC predicts the user intent within a domain such as PlayMusicIntent, BuyItemIntent, or MovieShowTimesIntent. And, NER recognize domain-specific named-entities such as artist name and song name for the Music domain, and item name and product type for the Shopping domain. 

In this paper, we investigate statistical model compression for NLU DC, IC, and  NER models. We use n-gram maximum entropy (MaxEnt) \cite{berger1996maximum} models for DC and IC, and n-gram conditional random fields models (CRF) \cite{Lafferty2001} for NER, but this work can be extended to any type of model with large number of features. We aim to reduce the large scale MaxEnt and CRF models memory footprint to enable local voice-assistants and decrease latency of loading skill NLU models in the cloud. We present two main techniques, parameter quantization and perfect hashing. We demonstrate these techniques' effectiveness with both empirical and theoretical justification. Also, we detail the tradeoffs of time, space, and predictive performance.






\section{Background and Related Work}

Various methods have been proposed to reduce the memory and CPU footprint of machine learning models for image classification \cite{Hubara2016, Han2015}, keyword spotting \cite{tucker2016model,sun2017compressed}, language models \cite{talbot2008randomized, lei2013accurate}, acoustic models \cite{wang2015small, mcgraw2016personalized}, and text classification \cite{joulin2016fasttext, ganchev2008small}. These methods fall into three classes. (i) Pre-processing methods -- these include classic dimensionality reduction techniques like principal component analysis, feature hashing \cite{Weinberger2009}, and random projection \cite{Dasgupta2000,Bingham2001} as well as deep autoencoders \cite{Hinton2006} and sparse autoencoders \cite{Ng2011}. (ii) Learning algorithm methods -- this is where the learning algorithm itself is programmed to produce a small model. Examples include L$1$-regularization, greedy step-wise feature selection, boosting of small-simplified models, and synaptic-pruning \cite{LeCun1989}. (iii) Post-processing methods -- these include methods such as parameter quantization \cite{Hubara2016} and data representation optimizations \cite{Han2015}.

Commonly, pre-processing and learning algorithm methods are already incorporated into the cloud model building process for a voice-assistant, so in this work, our efforts are primarily directed towards the post-processing methods. Parameter quantization has been shown to be effective for reducing memory footprint for both traditional and deep models. However, as far as we know perfect hashing has not been applied to MaxEnt and CRF compression, but only to language models \cite{talbot2008randomized}.

\section{Model Compression Approach}
\label{sec:techniques}

\subsection{Objective}
Our primary objective is to design algorithms which take large statistical NLU models and produce models which are equally predictive but have smaller memory footprint. This post-processing compression allows for reusing existing model building configurations and pipelines without maintaining separate ones for small-footprint models.




We evaluate the statistical model size reduction along three dimensions: \emph{time}, \emph{space}, and \emph{predictive performance}. Time refers to the computational runtime complexity to perform a prediction. Space is measured as the number of bits required to store the model in memory. We use the term predictive performance to refer to the evaluation metric of choice such as F1 score and accuracy.  Challenges arise in balancing the tradeoffs across these three dimensions as often improving on one will cost on the other two. For example, improving model prediction performance may require larger models and slower decoding time; while an effort to reduce decoding time may degrade predictive performance and increase the memory needed. Thus, it is important to find the best tradeoff for any given application. 


\subsection{Our Techniques}

We propose two techniques to perform statistical model compression \emph{quantization} and \emph{perfect hashing}. Individually, these approaches yield moderate model size reduction, but we combine them to achieve significant compression rates with minimal time and predictive performance tradeoffs. Before detailing the algorithms, we now briefly discuss a generalized model structure with accompanying notation. 

A machine learning model's memory footprint can be viewed principally as a large map from feature name to numeric weight. In NLU typically, there is an initially large universe $\featureuniverse$ of potentially active or relevant features (such as all English bigrams). Of those features, a subset $\featureset$, whose cardinality can be much smaller than that of $\featureuniverse$, are the relevant parameters chosen by the learning algorithm using feature selection methods. The relevant features and their corresponding weights are stored in the map while the irrelevant parameters have $0$ weight or are simply excluded from the model. For convenience we denote $\numentries = |\featureset|$ and assume all $0$ weight parameters are excluded from $\featureset$.

At runtime, to use the model to evaluate an instance, a set $\featuresubset \in \featureuniverse$  of parameters will be accessed. For MaxEnt and CRF, this instance parameter set will be small, $|S'| \ll |S|$.  Thus, for each instance only a relative small number of parameters will be required to make a prediction. For example, if $\featureuniverse$ is all English bigrams, $\featureset$ would be a smaller set of those bigrams which are useful features for the prediction task, and $\featuresubset$ would be those bigrams present in a single utterance. Hence $\featureset \cap \featuresubset$ are those parameters and weights needed to be accessed to predict on that instance.

If the parameter map is implemented as a hash map, the model memory footprint in total bits will be $O(\numentries \cdot (\stringlength + \weightsize))$ where $\numentries$ is the number of total parameters of the model, and $\stringlength$ and  $\weightsize$ are the sizes of the parameter name and weight respectively in bits.  The expected lookup time for a parameter is then $O(\stringlength + \weightsize)$, which is constant for bounded  $\stringlength$ and  $\weightsize$. Our goal is reduce this footprint while maintaining the lookup cost with little to no predictive performance degradation. 


\subsection{Quantization}
\label{sec:quantization}

Our initial step to model compression is model parameter quantization. To apply parameter quantization, we first choose a set of representative value cluster centers and then assign each parameter to its nearest cluster. When a parameter weight is accessed, its representative value is used in-place of the original value during the computation.  The advantage from a data storage perspective is that we now need only store the cluster identifier at each entry in our map instead of the full weight. Each weight in the map will be replaced by its cluster index which requires only $O(\log \numbins)$ bits where $\numbins$ is the number of clusters chosen.  Additionally, we must now store a small table of corresponding weights mapping each index to the representative cluster centers. And to predict a new instance we execute the computation by looking up the quantized index for each feature of $\featureset \cap \featuresubset$ then determining their quantized weights from the small table.  

In terms of space,  our parameter name to quantized index map is now reduced to a size of $O(\numentries (\stringlength + \log \numbins) )$ while our small table is of size $O(w \numbins)$ for a total size of $O(\numentries (\stringlength + \log \numbins) + w \numbins)$. In terms of runtime speed, the expected lookup time remains $O(\stringlength + \weightsize)$ with an additional cache miss due to the use of a second table. Using 256 bins requires only 8 bits per entry value to store the weights, as opposed to the 64 bits required for double precision or 32 bits for float precision.

For the predictive performance tradeoff, the two hyperparameters are the number of centroids $\numbins$ and the method for choosing the centers. Choosing the cluster centers for quantization can sometimes be a challenging task and many methods have been proposed. In the traditional \emph{linear quantization} the clusters are chosen by evenly partitioning the range between min and max weight values. We find that for our purpose, linear quantization yields adequate predictive performance results. The reason is that it rounds many small parameter values to zero, and preserves the larger weights that affect predictive performance. If the cluster centers were designed to follow the distribution of parameter values (peaky distribution around zero), this rounding effect would be smaller and the larger more important weights would also have less precision.



\subsection{Perfect Hashing}
Examining the total memory cost after quantization, we find that the $O(\numentries \stringlength)$ term dominates the memory footprint. However, at runtime, we can replace the full feature names set $\featureset$ using an elegant application of \emph{perfect hashing}. 

A perfect hash function maps our set $\featureset$ of $\numentries$ keys into $\hashbuckets$ buckets with no collisions and better yet a \emph{minimal perfect hash function} (MPHF) hashes our set $\featureset$ of $\numentries$ keys into $\numentries$ buckets with no collisions. If we had MPHF, then we just need to store an array of quantized indices and at runtime use the MPHF to index to the values of the parameters required for that instance. The challange is to find a hash function that achieves no collisions, is quick to evaluate, and requires little storage space. Here we describe a method which achieves $O(\numentries)$ expected space and  $O(\numentries)$ expected construction runtime, which is a variation of the method given in \cite{Limasset2017}.

Before describing the algorithm, we assume that we have access to a universal hash family from which we can draw pseudo-random hash functions, i.e. there is a set of hash functions $h_0, h_1, h_2, \dots$ where $h_i$ is the hash function with seed $i$ and each $h_i$ meets the simple uniform hashing assumption (SUHA). SUHA states each element hashed has an equal chance of being hashed to each bucket, meaning that $\prob{}{h_i(x)\modm m= \ell} = 1/\hashbuckets$ for all choices of $i$, $x$, and $\ell \in \left[0, \hashbuckets-1 \right]$. We also assume computing $h_i(x)$ is linear in the size of $x$. In practice, we implement this with a seeded MurmurHash \cite{Wikipedia2018}. With these assumptions in place, in Algorithm~\ref{mphf_construct} we outline the procedure for constructing a minimal perfect hash function from a set of keys.

\begin{algorithm}[h]

\caption{Minimal Perfect Hash Function  Construction}
\label{mphf_construct}

\begin{algorithmic}[1]
\State Set $\featureset_1 = \featureset$
\For{i $=$ 0,1,3,..}
\State Choose a hash function $h_i$. \label{}
\State Initialize a bit array $B_i$ of size $|\featureset_i|$ to zeros.
\State Hash all $x \in \featureset_i$ to range $\left[0, |\featureset_i| - 1\right]$.
\If {single entry of $\featureset_i$ gets hashed by $h_i$ to  position $j$}
\State Set $B_i[j] = 1$.
\EndIf
\State Set $\featureset_{i+1} = \left\{ x \in \featureset_i  \text{ where } B_i[j] \neq 1 \right\}$.
\If {$|\featureset_{i+1}| = 0$} 
\State break 
\EndIf
\EndFor

\end{algorithmic}

\end{algorithm}

Note that in Algorithm~\ref{mphf_construct}, we have a single unique $1$ set for each element of $\featureset$ through $B_1$, $B_2$, $\dots$. To find the hash of an element $x$ at runtime, we hash level by level until we hash to a $1$.  We then need to associate that $1$ with a unique index in range $[0, n - 1]$. This is done by viewing the bit arrays as one giant bit array $B = B_1 \oplus B_2 \oplus B_3 \oplus \cdots$ concatenating the arrays together and then assigning each $1$ to its \textit{rank}, i.e. the number of $1$'s preceding it in $B$.  This defines our minimal perfect hash function which we denote as $h^*$.

Concerning evaluation time, computing the rank can be done efficiently by using what are known as \textit{succinct data structures} \cite{Jacobson1988, Raman2002, Wikipedia2018a}. We break $B$ into chunks and store preaggregate rank sums computed at the chunk level. To find the rank of an index, we first look up the chunk level $1$'s count for the chunk containing the queried index. Then we linearly scan the containing chunk to compute the rank of the index inside the chunk. We return these two numbers added together. We can theoretically achieve constant time rank computation with a linear number of extra bits by having a multilevel chunking scheme (chunks within chunks). In practice, a simple one level scheme works well, especially since using bitwise operations during the linear scan of a chunk is particularly CPU cache efficient.

We now discuss how our minimum perfect hashing algorithm affects predictive performance. For every element $x \in S' \cap S$, the algorithm will return the correct index of that parameter weight. However, the lookup algorithm for $x \in S'\setminus S$ will either reach a $0$ at the bottom level bit array, in which case we know for certain the feature has a $0$ weight associated with it, or otherwise the $x$ will collide with another arbitrary parameter. This behavior is undesirable, and unless we explicitly store each key, which we are trying to avoid, it is impossible to guarantee no false-positives. However, we can reduce the false-positive rate at the cost of a few extra bits per entry. The idea is to store an extra array $F$ of entry ``fingerprints''. Given hash function $f$, we store $F[h^*(x)] = f(x) \modm (1/\epsilon)$ where $\epsilon$ is the desired false-positive rate. The fingerprint length will be $\log (1/\epsilon)$ bits in length and by SUHA the likelihood that two entries have matching fingerprints is $\epsilon$. Hence to lookup a weight we take the extra step of checking that its fingerprint matches. If the fingerprint does not match we can report a weight of 0 and that the parameter is not present in the model with 100\% certainty. Otherwise, we report the weight hashed to with $1 - \epsilon$ confidence.

From a space perspective, storing $h^*$ is small relative to the original $O(\stringlength \numentries)$ cost of storing each of the keys. It can be shown that $h^*$ will have size $O(\numentries)$ bits with high probability, but in order for the MPHF algorithm to be effective for our compression application, the hidden constant factors will need to be small. For the statistical models we deal with which commonly have $\numentries \geq 500,000$ parameters, the size of $h^*$ is less than $3.4 + \log (1/\epsilon)$ bits per entry.

We have addressed the predictive performance and space tradeoffs of applying the perfect hashing technique, so we last discuss the evaluation time tradeoff. We pay little extra in lookup time to use the perfect hashing algorithm. The expected evaluation time of $h^*(x)$ for $x \in U$ is $O(s + \log (1/\epsilon))$ with a constant number of expected cache misses. Table \ref{summary} below summarizes the tradeoffs discussed when applying the two techniques, quantization and perfect hashing, for compressing the models.

\begin{table}[h]
\centering
\caption{Summary of compressed models tradeoff}
\label{summary}
\setlength\extrarowheight{2pt}
{\footnotesize
\begin{tabular}{c|c|c|}
\cline{2-3}
                        & Normal         & Compressed        \\ \hline
\multicolumn{1}{|c|}{Parameter Access Time} & $O(\stringlength + \weightsize)$ & $O(\stringlength + \weightsize + \log (1/\epsilon))$ \\ \hline
\multicolumn{1}{|c|}{Total Space} & $O(n(\stringlength + \weightsize))$ & $O(\numentries \log(k/\epsilon) + w \numbins)$ \\ \hline
\multicolumn{1}{|c|}{\shortstack{Predictive Performance \\ Impact}} & \shortstack{baseline \\ \phantom{a}} & \addstackgap{\shortstack{granularity loss, \\ $\epsilon$ false-positive rate}} \\ \hline
\end{tabular}
}
\end{table}

\section{Experimental Results}
\label{sec:results}

In this section we present experimental results using our compression techniques. We apply them to large scale cloud NLU models from six domains suitable for local voice-assistants, such as temperature control and navigation. All of these models have feature counts over 500,000 and many have beyond several million. We first discuss the model size reduction and then effect on the predictive performance.

Note that for skill NLU models the results are similar and not provided here.

\subsection{Compression}
\label{sec:results_compression}
The compression results are given in Table \ref{table:compressed_sizes}. After applying the model size reduction techniques, we see significant compression rates compared to the normal statistical models. We used hyperparameters of $k = 256 \Rightarrow \log k  = 1 \text{ byte}$ and false-positive rate $\epsilon = 0.0001$. We had experimented with varying $k$ but found $k = 256$ a desirable choice because it gave us adequate predictive performance and is programmatically convenient since each cluster index can be stored with a whole byte. We achieve a significant 14.25-fold memory footprint reduction (567.2 MB compared to 39.8MB) and for some models have a compression ratio as high as a 31.5 (Domain 1 IC). 

The MaxEnt DC total compression rate is lower that the MaxEnt IC models, 12.3-fold vs. 24.7 fold. The reason is that for the normal DC we use the feature hashing trick during training. Thus, instead of storing a map from string feature name to feature id, we store a 32-bit integer hash to feature id map. Since our DC models have millions of parameters, this integer map takes significant memory, and using our approach reduces the memory requirement for each key from 32-bit value to less than $3.4 + \log (1/\epsilon)$.

The CRF NER total compression rate is around 10.8-fold, which is lower than MaxEnt. The reason is that CRF models have greater structural complexity than MaxEnt, and we need to maintain additional information on state transitions and state observation that we do not quantize and hash.

Without fingerprinting ($\epsilon = 1$), we obtain 25.3-fold memory footprint reduction (567.2 MB compared to 22.4MB). This is around 43\% lower compared to $\epsilon = 0.0001$. From Table \ref{table:compressed_sizes}, we note that for IC and DC models the fingerprints consume more than half of the memory. However, without fingerprinting the false-positives from parameter access affect predictive performance which is described in the next section.



\begin{table*}[!ht]
\centering
{\footnotesize
\caption{NLU statistical models sizes in megabytes. Normal vs. compressed ($\epsilon = 0.0001$) vs. compressed* ($\epsilon = 1$)}
\label{table:compressed_sizes}
\setlength\extrarowheight{1.5pt}
\begin{tabular}{|c|ccc|ccc|ccc|ccc|} 
\hline                                                                                                                                                                                                                                         
          & \multicolumn{3}{c|}{DC}                                          & \multicolumn{3}{c|}{IC}                                          & \multicolumn{3}{c|}{NER}                                         & \multicolumn{3}{c|}{All}                                       \\
          & \multicolumn{1}{c}{Normal} & \multicolumn{1}{c}{Comp.} & Comp.* & \multicolumn{1}{c}{Normal} & \multicolumn{1}{c}{Comp.} & Comp.* & \multicolumn{1}{c}{Normal} & \multicolumn{1}{c}{Comp.} & Comp.* & \multicolumn{1}{c}{Normal} & \multicolumn{1}{c}{Comp.} & Comp* \\ \specialrule{1.0pt}{0.5pt}{0.5pt}
Domain 1  & 27.2                       & 2.2                        & 0.9    & 42.                        & 1.3                        & 0.4    & 14.0                       & 1.5                        & 1.2    & 83.2                       & 5.0                        & 2.5    \\ 
Domain 2  & 25.8                       & 2.1                        & 0.9    & 65.4                       & 1.4                        & 0.5    & 86.9                       & 7.6                        & 5.5    & 178.1                      & 11.1                       & 6.9    \\ 
Domain 3  & 13.1                       & 1.1                        & 0.4    & 0.6                        & 0.1                        & 0.002  & 1.0                        & 0.2                        & 0.1    & 14.7                       & 1.4                        & 0.5    \\ 
Domain 4  & 10.9                       & 0.9                        & 0.4    & 9.5                        & 0.6                        & 0.2    & 2.5                        & 0.4                        & 0.3    & 22.9                       & 1.9                        & 0.9    \\ 
Domain 5  & 25.3                       & 2.0                        & 0.9    & 25.7                       & 1.3                        & 0.5    & 84.7                       & 7.4                        & 5.6    & 135.7                      & 10.7                       & 7.0    \\ 
Domain 6  & 51.5                       & 4.1                        & 1.7    & 64.8                       & 3.7                        & 1.3    & 16.4                       & 1.9                        & 1.6    & 132.6                      & 9.7                        & 4.6    \\ \hline
Total     & 153.7                      & 12.4                       & 5.2    & 208.0                      & 8.4                        & 2.9    & 205.5                      & 19.0                       & 14.3   & \textbf{567.2}                      & \textbf{39.8}                       & \textbf{22.4}   \\ \hline
\end{tabular}
}
\end{table*}

\subsection{Predictive Performance}
\label{sec:results_predictive_performance}


We evaluate model performance on two large test datasets with hundreds of thousand annotated utterances: 
\begin{itemize}[topsep=2px, leftmargin=15pt]
\item Supported Domains (SD) Test set: Contains utterances from the six local supported domains.
\item Out of Domain (OOD) Test set: This includes utterances that do not map to any intent or background noise.
\end{itemize}
We use the following evaluation metrics:
\begin{itemize}[topsep=2px, leftmargin=15pt]
\item Slot Error Rate (SER) \cite{makhoul1999performance} is a slot level metric that evaluates the over all predictive performance of the models. SER is defined as the ratio of the number of slot prediction errors to the total number of reference slots. Errors could be insertions, substitutions and deletions. Intent misrecognitions are considered substitutions.  
\item Intent Classification Error Rate (ICER) utterance level metric. ICER is defined as the ratio of the number of intent misclassifications to the total number of utterances. 
\item F-ICER is a balanced ICER metric that considers both precision and recall. We compute it as 1 - F1 score.
\item Rejection Rate is defined as the percentage of utterances with scores below a set threshold. Below threshold utterances are rejected by the system.
\end{itemize}
An NLU system ideally has a low SER and ICER/F-ICER on the SD test set indicating good model predictive performance and a high rejection rate on the OOD test set indicating that out of domain utterances are rejected. 

Table \ref{tab:quantization_res_overall} details the overall performance measures and Table \ref{tab:quantization_res_domain} details the per domain performance measures. The results are percentage relative compared to the normal models, as we are unable to disclose absolute numbers. 

As shown from the results, our compressed models perform almost as well as our baseline models with acceptable overall relative error increases of +0.86\% in SER and +0.26\% in ICER. The per domain compressed results show small F-ICER increase of less than +1\% except Domain 3 with +1.58\%. The reason is that Domain 3 IC model has a small number of prominent features and false-positive on important features are more common.

The ultra compressed models without fingerprinting ($\epsilon = 1$) have overall relative error increases of +2.20\% in SER and +3.14\% in ICER. The per domain relative error increases are around +2\% to +3\% F-ICER for most domains. The error rates of the ultra compressed models are higher than the compressed models with fingerprinting. However, for the relative error increase of around +1 to +2\%, we obtain total 25.3-fold memory reduction compared to 14.25-fold. Thus, compression without fingerprinting could be a viable option depending on the memory constraints and predictive performance requirements.

Note that the reason why using no fingerprinting performs adequately is because when we get a false-positive, it is not as if an adversarial index is hashed and the system is guaranteed to make an incorrect prediction. Rather, when a false-positive is realized, we actually are hashing to a random existing model parameter each with equal probability. Since majority of our parameters are close to zero, the false-positives add small amount of noise to the predictions.

\begin{table}[t]
\centering
{\footnotesize
\caption{Overall predictive performance measures for NLU statistical models.} 
\label{tab:quantization_res_overall}
\begin{tabular}{@{}ccccc@{}}
\toprule
\multirow{2}{*}{Model}  & \multicolumn{2}{c}{SD Dataset}        & OOD Dataset  \\ 
                                                                                                                                     & SER & ICER & Rejection Rate                \\ \midrule
Compressed                  & +0.86\%                     &    +0.26\%                       & -0.08\%                 \\
Compressed*                 & +2.20\%                     &    +3.14\%                       & -0.72\%                 \\ \bottomrule
\end{tabular}
}
\end{table}


\begin{table}[t]
\centering
{\footnotesize
\caption{Domain predictive performance measures for NLU statistical models on the SD test set.} 
\label{tab:quantization_res_domain}
\begin{tabular}{@{}ccccc@{}}
\toprule
\multirow{2}{*}{Domain}  & \multicolumn{1}{c}{Compressed}        & \multicolumn{1}{c}{Compressed*}  \\ 
                           & F-ICER 					 & F-ICER        \\ \midrule
Domain 1                  &    +0.01\%       & +2.33\%                 \\ 
Domain 2                   &    +0.74\%         & +2.36\%                 \\ 
Domain 3                 &    +1.58\%       & +8.94\%                 \\ 
Domain 4                  &    +0.01\%       & +1.75\%                 \\ 
Domain 5                  &    +0.32\%   & +1.59\%                 \\ 
Domain 6                  &    +0.20\% 	& +3.15\%                 \\ 
\bottomrule 
\end{tabular}
}
\end{table}

\section{Conclusion}

In this paper we presented approaches to reduce the memory footprint of NLU statistical models to work on resource-constrained embedded systems, and decrease latency of loading skill NLU models. We demonstrated the effectiveness of our techniques in reducing memory footprint while addressing the the tradeoffs of time, space, and predictive performance. We observed the methods sacrifice minimally in terms of model evaluation time and predictive performance for the substantial compression gains observed. It would be interesting to go beyond the results of Section \ref{sec:quantization} to see if there is a better quantization scheme for our models. 

\bibliographystyle{IEEEtran}

\bibliography{references}


\end{document}